\documentclass{article}

\twocolumn
\usepackage[utf8]{inputenc}
\usepackage{hyperref}
\usepackage[a4paper]{geometry}
\usepackage{graphicx}
\usepackage{xurl}
\usepackage[numbers]{natbib}
\newcommand\blfootnote[1]{%
  \begingroup
  \renewcommand\thefootnote{}\footnote{#1}%
  \addtocounter{footnote}{-1}%
  \endgroup
}

\begin{document}

\title{\textbf{nuReality: A VR environment for research of pedestrian and autonomous vehicle interactions}}
\author{Paul Schmitt, Nicholas Britten, JiHyun Jeong, Amelia Coffey,  \\Kevin Clark, Shweta Sunil Kothawade, Elena Corina Grigore, \\Adam Khaw, Christopher Konopka, Linh Pham, Kim Ryan,\\ Christopher Schmitt, Aryaman Pandya, Emilio Frazzoli} 

\date{December 10, 2021}
\maketitle


\begin{abstract}
\emph{}
{\small 
We present nuReality, a virtual reality ``VR''  environment designed to test the efficacy of vehicular behaviors to communicate intent during interactions between autonomous vehicles ``AVs'' and pedestrians at urban intersections. In this project we focus on expressive behaviors as a means for pedestrians to readily recognize the underlying intent of the AV’s movements. VR is an ideal tool to use to test these situations as it can be immersive and place subjects into these potentially dangerous scenarios without risk. nuReality provides a novel and immersive virtual reality environment that includes numerous visual details (road and building texturing, parked cars, swaying tree limbs) as well as auditory details (birds chirping, cars honking in the distance, people talking).  In these files we present the nuReality environment, its 10 unique vehicle behavior scenarios, and the Unreal Engine and Autodesk Maya source files for each scenario. The files are publicly released as open source at \href{http://www.nuReality.org}{www.nuReality.org}, to support the academic community studying the critical AV-pedestrian interaction.\blfootnote{This research was supported by Motional, Boston, MA 02210, USA.}
\blfootnote{Please direct all correspondence to nuReality@motional.com} 

}
\end{abstract}

\section{Introduction}

Motional sees effective communication between AVs and pedestrians as a key challenge to acceptance and adoption within society. In order for a pedestrian to feel safe, the communication between the pedestrian and AV should be simple and and easy to understand. Ideally the AV would clearly and intuitively indicate its intentions, as in this environment, to stop and let a pedestrian cross the road. In order to be deemed intuitive, the communication should follow patterns that familiar to the pedestrian.  In this way, pedestrians across different communities would not need to pause, view, and learn a new set of signals.

In these files we present nuReality, a VR environment designed to address some of the challenges we encountered with existing simulation tools to test the efficacy of vehicular behaviors to communicate intent during interactions between AVs and pedestrians at urban intersections. nuReality was designed specifically to recreate a life-like visualization from the perspective of a pedestrian attempting to cross the street. Figure 1 shows a panoramic view of the environment. The following sections detail the contents of \href{http://www.nuReality.org}{nuReality} and how they can be leveraged to support AV research.

\begin{figure*}[ht]
\includegraphics[width=\linewidth]{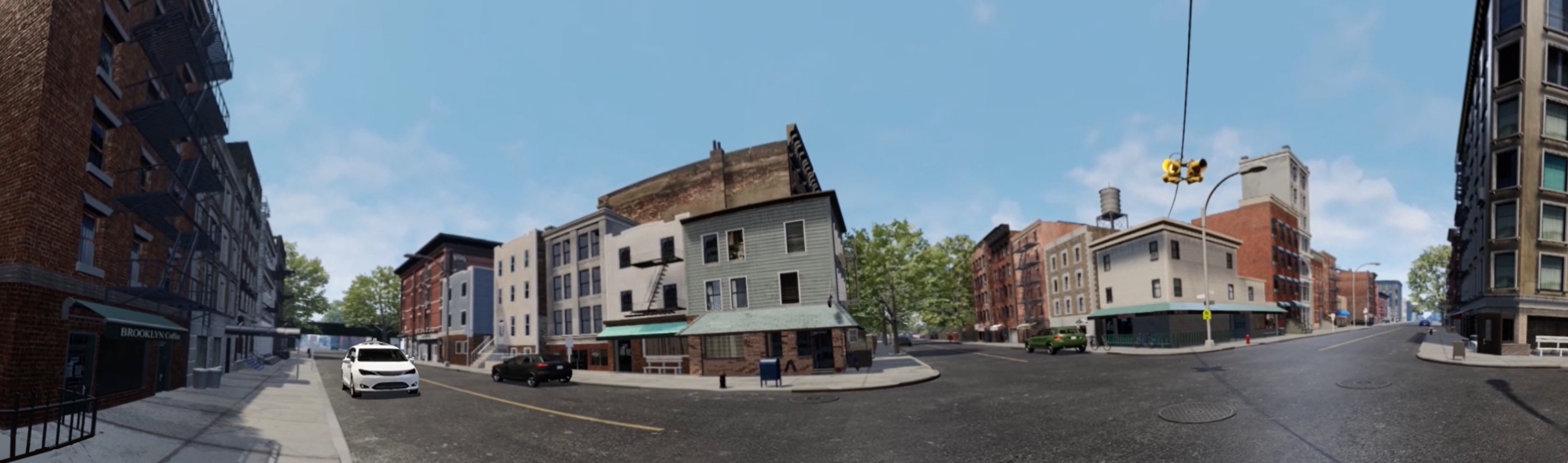}
\centering
\caption{Panorama of the virtual environment from the first-person perspective of a pedestrian about to cross the street as an AV approaches}
\end{figure*}

\section{Content of the Files}
The virtual reality environment is modeled after an urban 4-way intersection. There is no clear pedestrian crossing zone (zebra stripes), no stop signs, no stoplights, or other elements to indicate that the car lawfully would have to come to a complete stop. While there is a flashing yellow light above the intersection and a yellow pedestrian warning sign can be seen on the diagonally opposite street corner, these traffic signals are only cautionary and the pedestrian is uncertain if the vehicle will stop and allow them to cross the street. Significant care was taken in creating a realistic environment, as enhanced “place illusion” and “plausibility” would elicit higher fidelity of participant reactions \citet{Slater}. Numerous visual details (including but not limited to road and building texturing, parked cars and swaying tree limbs) as well as auditory details (birds chirping, cars, people talking) were included in order to achieve spatial presence within the virtual environment \citet{Wirth}. All scenarios were crafted initially using Autodesk Maya, brought into Unreal Engine for additional editing, and then compiled into new applications.

Two vehicle models are present in the files: a conventional human driven vehicle and a novel Autonomous Vehicle without a human operator. Both vehicles were modeled after a white 2019 Chrysler Pacifica, however the AV model has some notable differences. The AV model includes both side-mirror and roof mounted lidar and radar sensors and has no visible occupants. The human driven model includes a male driver who looks straight ahead and remains motionless during the interaction. 

	The following briefly describes each of the scenarios available within the \href{http://www.nuReality.org}{nuReality} files. 
{\small 
\begin{itemize}

\item Orientation: \textbf{ No vehicle.} Allows the user the get familiarized to the environment
\item Practice: \textbf{Human driver baseline vehicle.} Followed a deceleration curve modeled
after human drivers stopping at a stop controlled intersection in a natural setting
\item Scene 1: \textbf{Human driver baseline vehicle.} Followed a deceleration curve modeled after human drivers stopping at a 
stop-controlled intersection in a natural setting. See Figure 2
\item Scene 2: \textbf{AV baseline.} Followed the same deceleration curve as the Practice scenario and Scene 1. The vehicle is equipped with the AV sensor suite. See Figure 3
\item Scene 3: \textbf{AV with a Light Bar.} As the vehicle is yielding, a light bar mounted at the top of the windshield uses cyan colored LEDs to display a sweeping motion from the center to the edge bar. See Figure 4
\item Scene 4: \textbf{AV with Expressive Deceleration.} The behavior follows the animation principles of exaggeration and ease in/ ease out to make the deceleration less robotic. The AV exaggerates its motion by initiating the braking sequence 10 meters earlier and achieving a peak deceleration value 10\% greater than the baseline scenario. 
\item Scene 5: \textbf{AV Stops Short} The AV yields 5 meters farther away from the pedestrian than in the other conditions, coming to an early stop. See Figure 5
\item Scene 6: \textbf{AV uses Explicit Sound.} Exaggerated brake noises and low engine RPM sounds that increases in volume as the car approaches closer to the pedestrian were ad-ded to this scenario
\item Scene 7: \textbf{AV uses Expressive Nose Dive and Explicit Sound.} As the vehicle comes to a stop the nose is dipped down 1 inch the tail raises 1 inch before leveling out. This action is combined with the exaggerated sounds from Scene 6
\item Scene 8: \textbf{Human Driver Doesn’t Stop.} Appears to slow as it approaches the intersection but continues through without stopping
\item Scene 9: \textbf{AV Doesn’t Stop.} Appears to slow as it approaches the intersection but continues through without stopping
\item Scene 10: \textbf{AV with Expressive Nose Dive.} As the vehicle comes to a stop the nose is dipped down 1 inch the tail raises 1 inch before leveling out

\end{itemize}
}
\begin{figure}[h]
\includegraphics[width=\linewidth]{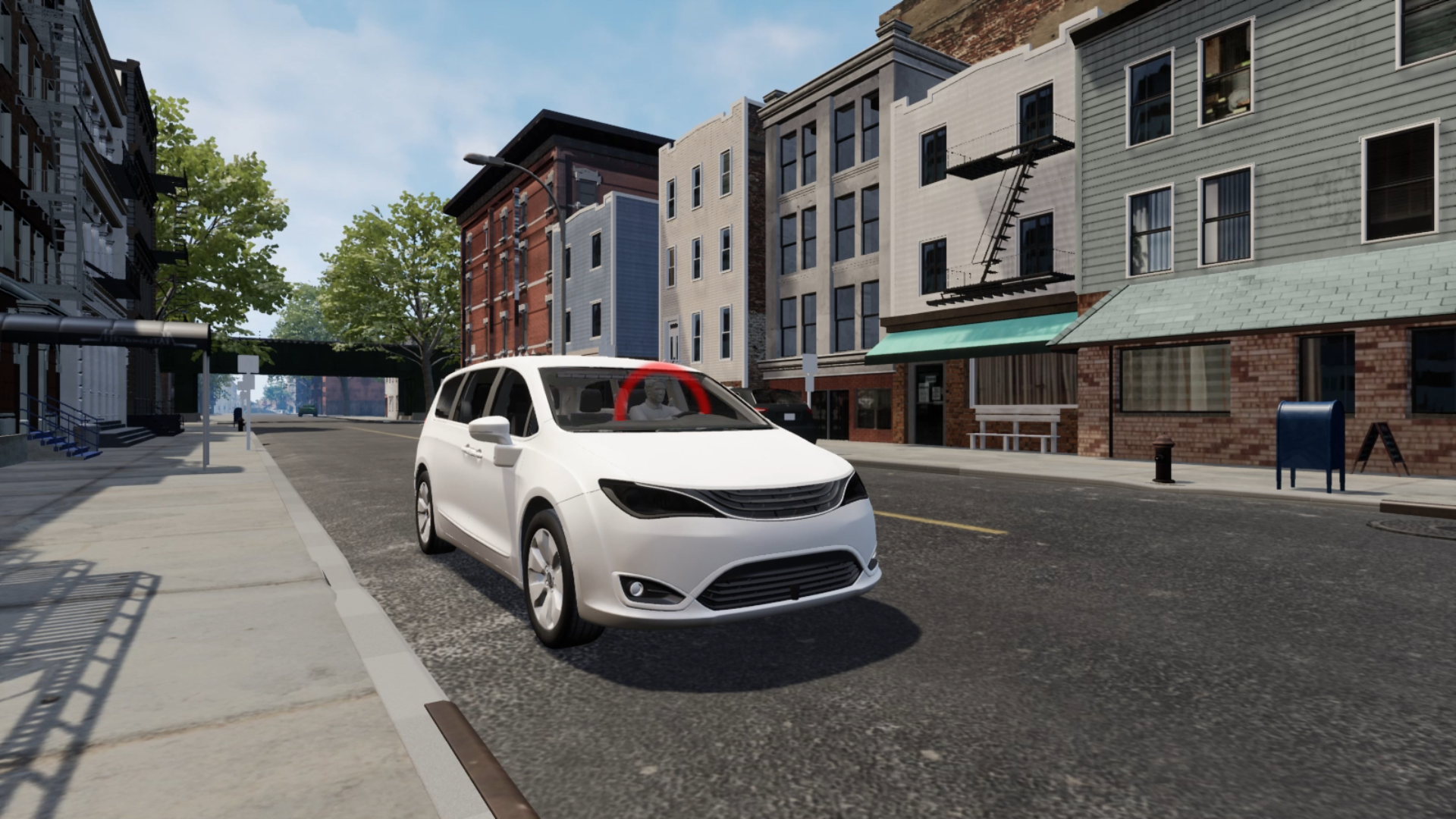}
\caption{Traditional vehicle  with  a  human  in  the  driver’s  seat}
\end{figure}

\begin{figure}[h]
\includegraphics[width=\linewidth]{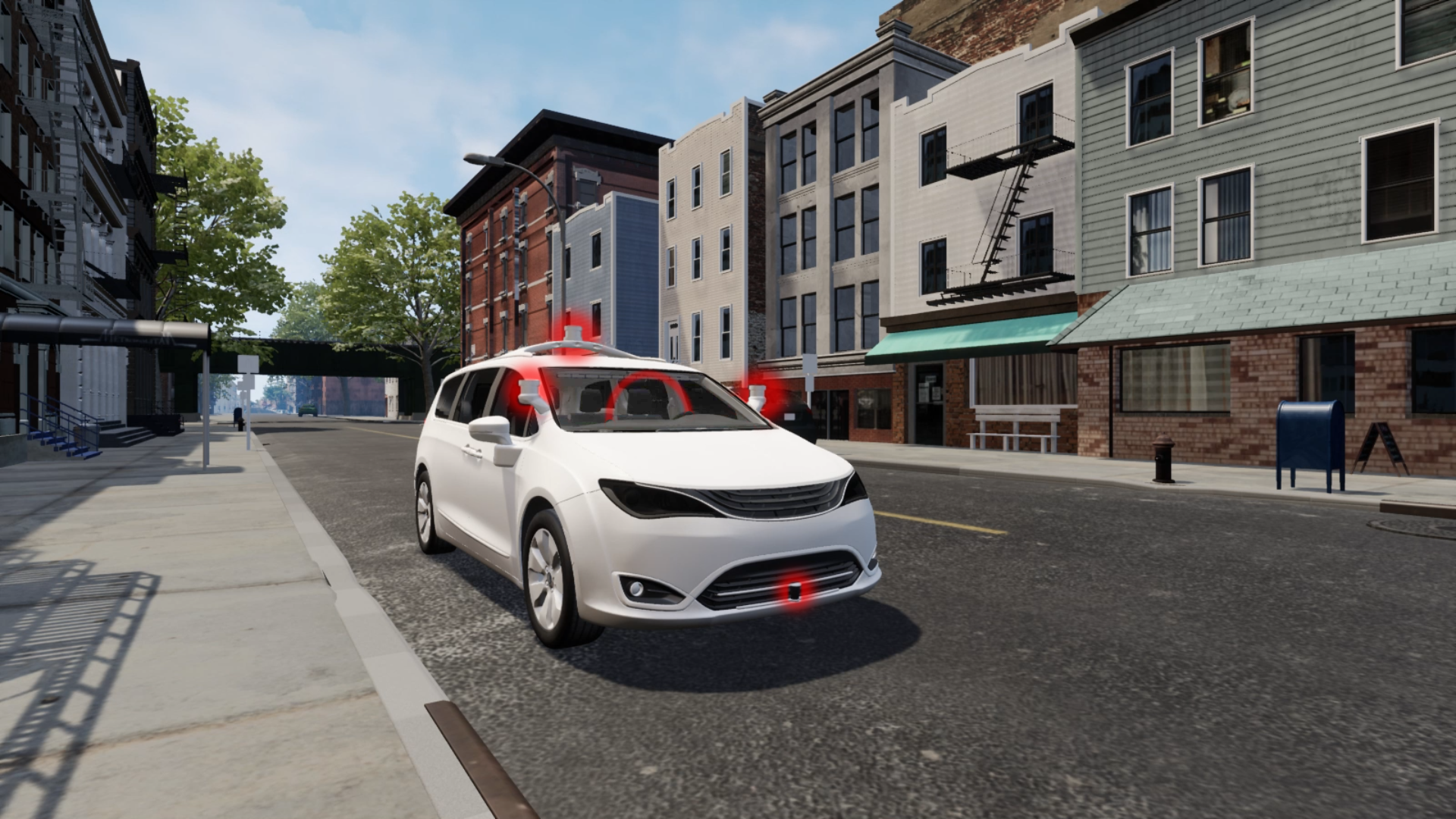}
\caption{Autonomous vehicle  with  visible LIDAR sensors and no human driver}
\end{figure}

\begin{figure}[h]
\includegraphics[width=\linewidth]{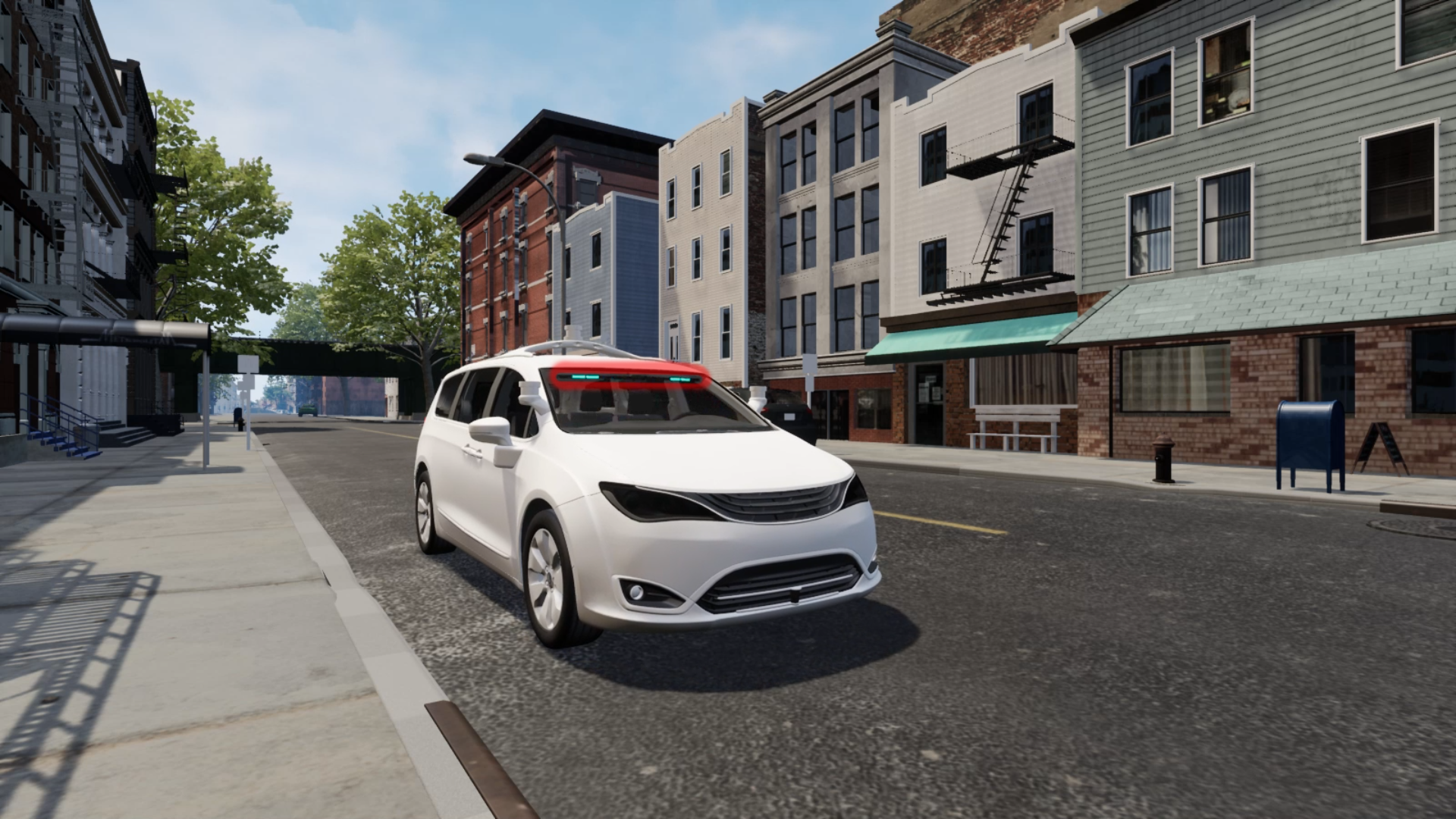}
\caption{Autonomous vehicle with cyan  LED lights that move in a sweeping motion}
\end{figure}

\begin{figure}[h]
\includegraphics[width=\linewidth]{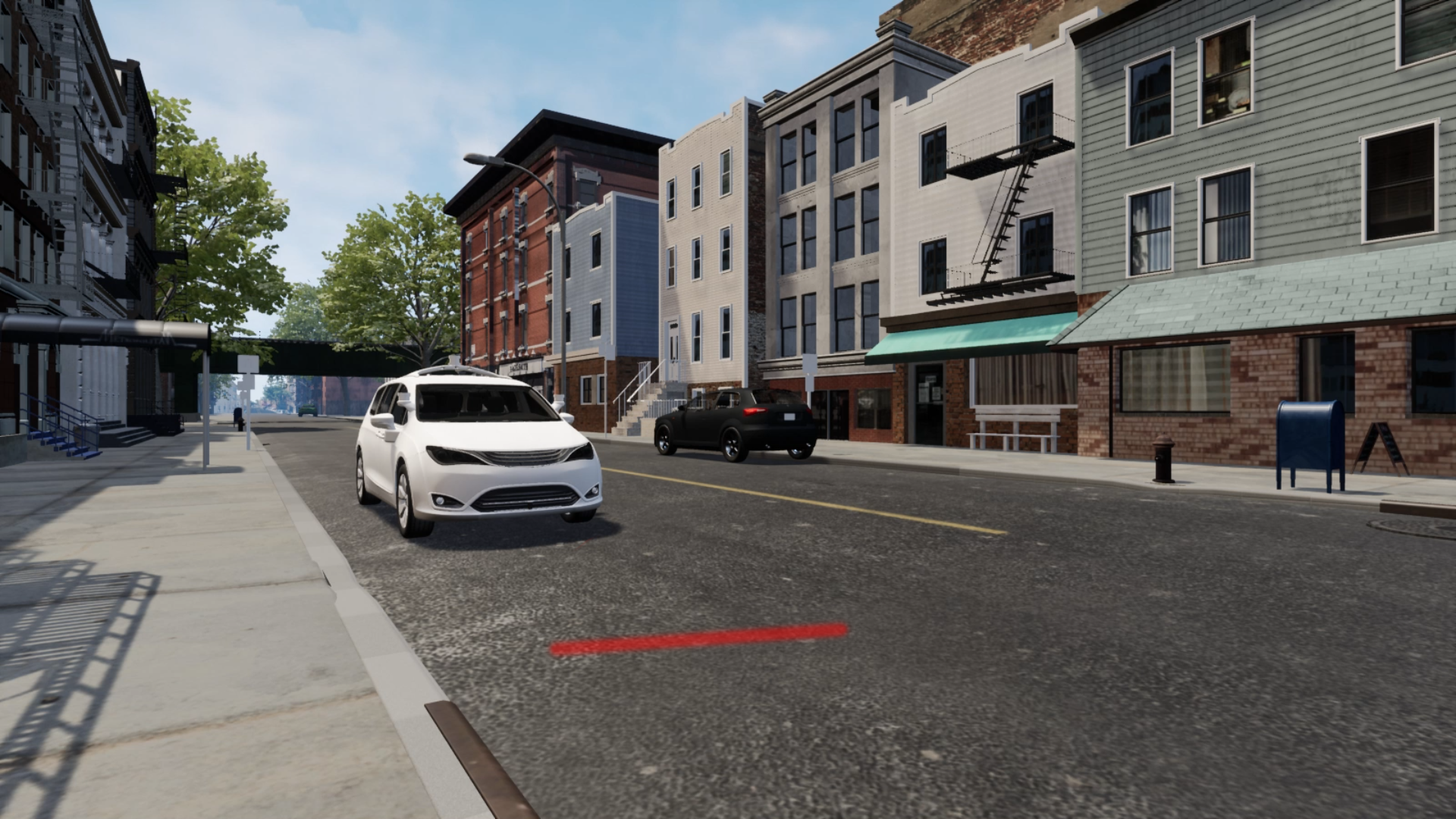}
\caption{Autonomous vehicle stops one car's length farther away from the pedestrian’s crossing point}
\end{figure}

\section{How nuReality Can Support AV Research}

The nuReality files are designed as a base to support experimentation and further research into expressive robotics, autonomous vehicles, pedestrian interactions, and related areas. Further, they are designed to overcome some of the challenges of existing simulation tools. First, nuReality was designed in Virtual Reality over other 3D alternatives in order to give users a truly immersive experience. A few advantages of nuReality over commonly used AV simulation tools are realism and HRI tailoring. Since our main goal was to foster high-fidelity reactions from trial participants, we worked with a VR studio to create a life-like environment.

As encounters with AVs become more frequent in people’s everyday lives, challenges of communication and inferring intention must be solved before driverless vehicles’ benefits of safety and accessibility can be appreciated. 
Trust and understanding in human-AV interactions is a precursor to widespread acceptance and adoption.

These challenges are bigger than a single company. The larger research community is instrumental to uncovering solutions for pedestrian interactions with driverless vehicles. That is why Motional has decided to use a VR platform and make these files open source. Over the past few years VR technology and systems have become more commercially available. The benefits of using VR technology lie not only in experimental control, reproducibility, and ecological validity; but also in accessibility \citet{Xueni}. While VR has been used in AV-pedestrian research before \citet{Schmidt}, we hope that nuReality will act as a catalyst in its adoption as an accessible means of experimentation. 

The contents of nuReality are as described in section 3, however we leave it up to the broader research community to build upon that foundation. Access to the Maya and Unreal source files gives researchers the ability to modify the given scenarios as well as the overarching environment. This opens up numerous possibilities including but not limited to developing scenarios with novel expressive behaviors, fine tuning the existing scenarios, and changing the environment to better suit different regions of the world. 

\begin{figure}[h]
\includegraphics[width=\linewidth]{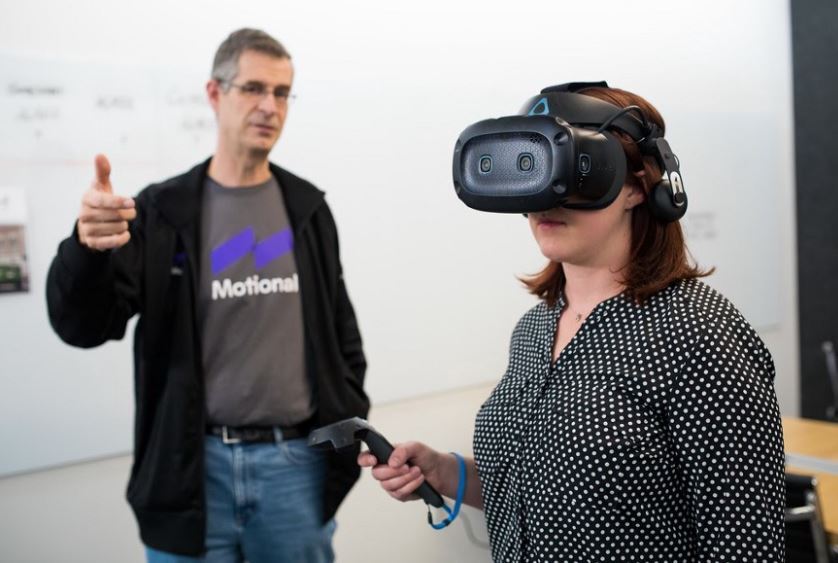}
\caption{Example of a participant using the VR equipment while receiving instructions}
\end{figure}

\section{Related Work}

Expressive (or Predictable, Legible) Robotics \citet{legibilty} is an exciting area of research that leverages principles from a variety of fields, ex. human factors \citet{human} \citet{Laura} and animation \citet{12}, to generate behaviors that enable an observer to quickly and confidently infer correctly the vehicle's  goal or to match behaviors to what an observer would expect from a human driver. 

Our work builds on the contributions from a small yet important community of researchers working on the challenges around AV communication. Anca Dragan’s Legible Robotics \citet{Dragan} and Andrea Thomaz’ Socially Intelligent Robots \citet{Thomaz} have been applied successfully to humanoid and mobile robots. Meanwhile, research such as InterACT \citet{interACT} and work at the University of Leeds \citet{UL} and Technische Ünive-rsität Münichen \citet{TUM}, is advancing our understanding of pedestrians' behaviors, needs, feelings, and mental flows. Solutions involving different technologies are also arising in order to fill in this communications gap. Much research has been performed on electronic Human Machine Interfaces (e-HMIs) to enable communication between the AV and pedestrian \citet{Fridman}. Many solutions, such as the Ford/VTTI light bar \citet{Ford} and Volvo 360 light ring \citet{Volvo}, utilize a combination of light patterns to communicate the vehicle’s intent. Others have also experimented with using sounds and simulated eyes \citet{Jaguar}. 

The scenarios in these files were designed specifically based on the principles formulated by this sphere of research.

\section{Conclusion and Outlook}
We present nuReality,  a virtual reality (VR) environment designed to test the efficacy of expressive behaviors to communicate vehicle intentions during interactions between AVs and pedestrians at urban intersections. We focus, in this project, on expressive behaviors as a means for pedestrians to readily recognize the underlying intent of an AV’s movements. The result of these files is a base off of which others can continue testing and building that will push research towards developing future generations of AVs that  
can effectively communicate with humans in complex situations. This communication will build trust between humans and AVs, help humans be more confident in their decision making around AV interactions, and make AV-human interactions safer.

\bibliographystyle{IEEEtranN}
\bibliography{reference}

\end{document}